\documentclass[11pt]{article}

\usepackage[final]{acl}

\usepackage{times}
\usepackage{latexsym}

\usepackage[T1]{fontenc}

\usepackage[utf8]{inputenc}

\usepackage{microtype}

\usepackage{inconsolata}
\usepackage{enumitem}

\usepackage{graphicx}

\providecommand{\us}{\_\allowbreak{}}

\usepackage{hyperref}
\usepackage{url}

\usepackage{times}
\usepackage{blindtext}
\usepackage{subcaption}
\usepackage{multirow}
\usepackage{multicol}
\usepackage{booktabs}
\usepackage{tikz}
\usepackage{pgfplots}
\usepackage{listings}
\usepackage{tcolorbox}
\usepackage{tabularx}
\usepackage[table]{xcolor}
\usepackage{pifont}
\usepackage{amsmath}
\usepackage{todonotes}
\tcbuselibrary{skins, breakable}

\newtcolorbox{promptbox}[1]{%
  enhanced,
  breakable,
  colback=gray!5,
  colframe=gray!60,
  boxrule=0.5pt,
  left=6pt, right=6pt, top=6pt, bottom=6pt,
  fonttitle=\bfseries\small,
  title={#1},
  coltitle=black,
  attach boxed title to top left={yshift=-2mm, xshift=4mm},
  boxed title style={
    colback=gray!5,
    colframe=gray!60,
    boxrule=0.5pt,
    left=3pt, right=3pt, top=1pt, bottom=1pt
  }
}

\definecolor{accent}{HTML}{1F4E79}
\definecolor{accent2}{HTML}{2E75B6}
\definecolor{lightbg}{HTML}{EAF1FB}
\definecolor{capturebg}{HTML}{F2F7FD}
\definecolor{greytext}{HTML}{595959}

\newtcolorbox{warnbox}{
  colback=lightbg,
  colframe=accent2,
  leftrule=3pt, toprule=0pt, bottomrule=0pt, rightrule=0pt,
  arc=0pt, outer arc=0pt,
  left=8pt, right=6pt, top=5pt, bottom=5pt,
  breakable
}

\pgfplotsset{compat=1.17}
\usepgfplotslibrary{groupplots}

\title{\textsc{Athaar}: A 
Medieval Scientific Arabic Manuscript Dataset for Handwritten Text Recognition}

\author{
 \textbf{Ismail El Jamiy\textsuperscript{1,2,3}},
 \textbf{Khadija Msalek\textsuperscript{2}},
 \textbf{Guillaume Loizelet\textsuperscript{2}},
 \textbf{Farah Benamara\textsuperscript{1}} \\
\textbf{Nabil Hathout\textsuperscript{3}} \\
 \textsuperscript{1}IRIT, Université de Toulouse, France \\
 \textsuperscript{2}IMT, Université de Toulouse, France 
 \\
 \textsuperscript{3}CLLE, Université de Toulouse Jean-Jaurès, France \\
 \small{\textbf{Correspondence:} \href{mailto:email@domain}{ismail.el-jamiy@irit.fr}}
}

\begin{document}
\maketitle
\begin{abstract}
Despite recent progress in handwritten text recognition, medieval Arabic scientific manuscripts remain an open challenge for current vision-language models (VLMs) due to 
cursive ligatures, variable diacritization, and more importantly the lack of training data.
This paper aims to address this gap, by proposing  the first annotation guidelines dedicated to the transcription of scientific Arabic manuscripts 
inspired by 
textual criticism, a well-established approach to historical text editing. We then propose,  \textbf{\textsc{Athaar}}, a new dataset manually annotated following this guideline. Unlike previous resources in the field, \textsc{Athaar} 
varies both witnesses of the same text, calligraphic styles, and content to deal with texts but also numerals. Relying on state-of-the-art VLMs, we finally design a set of experiments to study models transfer across scripts, chapters, and treatises. Our findings  highlight the limitations of current systems when faced with the specificities of ancient scientific texts. 
\end{abstract}

\section{Introduction}
\label{sec:introduction}
The preservation and analysis of historical documents are crucial to
understanding the evolution of human societies. Worldwide, millions of
manuscripts written in various languages, such as ancient Greek, Classical Chinese and  Latin, remain untapped in archives and libraries. These
documents constitute invaluable primary sources containing data on various
disciplines like philosophy, science, and literature. However, access to
this knowledge remains limited because, while large-scale digitization
allows for the creation of images, their textual content often remains
inaccessible to modern methods of indexing and computational processing \citep{chan_hatformer_2025}.

To address this need, the scientific community has developed tools for
Handwritten Text Recognition (HTR). Progress has been made for certain
languages, such as Classical Chinese \citep{zhao_fuxi_2025}, Medieval
Latin \citep{koch-etal-2023-tailored} or Old French
\citep{clerice_pre-editorial_2026}, demonstrating the ability of
artificial intelligence to transcribe and structure complex historical
corpora. For Arabic, despite manuscript collections being estimated at over three million
volumes (one of the most extensive written traditions in history), automatic recognition of handwritten texts has lagged behind,
 partly due to the lack of training data \cite{sommerschield_machine_2023}. Existing corpora 
 have been compiled from a variety of textual sources (newspapers, letters, manuscripts), such as Muharaf \cite{saeed_muharaf_2024}, or through the collection of modern handwritten texts produced by volunteers, such as KHATT \cite{mahmoud_khatt_2012}.

As our study focuses on historical manuscripts,  a closer analysis of existing HTR datasets  reveals three main limitations for our use case: 
(1) \textit{Under-representation of the scientific domain.} Available resources are composed of  religious  (such as Agapet \citep{ibrahim_agapet_2024}, and Historical Arabic HTR \cite{NAJAM2024110813}), literary texts (e.g., Iskandar \cite{iskandar_2024},  RASAM \cite{vidalgorene:hal-03971691}). 
These datasets are therefore not suited to deal with texts such as those found in the exact sciences, where  domain-specific terminology, numerals and geometrical
objects (e.g., angles) 
compound the challenges of automatic recognition.
(2) \textit{Lack of annotation guidelines.} Annotation conventions guiding manual transcription are rarely
documented and none report inter-annotators  agreement raising  questions about their reliability for evalutaing, training or fine-tuning HTR models. 
(3) \textit{Limited diversity in scripts and textual witnesses.}  Variability in writing styles across manuscripts of the same text is limited. Indeed, to ensure stable learning and avoid biases (such as overfitting to a single scribe), HTR models must be exposed to a variety of handwriting styles for the same content.




This paper aims to bridge these gaps, by proposing
 \textsc{Athaar}, the first  medieval scientific Arabic manuscript dataset for
handwritten text recognition. Unlike previous resources, \textbf{\textsc{Athaar}} includes   manuscripts from  major intellectual figures of the  Islamic Golden Age, while varying both witnesses of the same text, calligraphic styles, and content to deal with texts but also numerals. It is built around \textit{al-Qānūn al-Mas'ūdī}, an
astronomical treatise composed by al-Bīrūnī around 1035 that survives in
many manuscript copies. It also includes other scientific treatises from the same period covering astronomy and mathematics. Our contributions are:

 \textbf{(a) An annotation guideline  dedicated to the transcription of Arabic manuscripts grounded in textual criticism,}  a well-established approach to historical text editing
    that records in a critical edition \textit{variant reading}, i.e., 
 any difference in wording, spelling, or form between two or more
witnesses of the same text at the same location \citep{Maas_1958,
west_textual_1973}. As far as we know, this is the first publicly released guide
    covering textual content, numerals and geometrical objects. Although its primary focus is on scientific content, the methodology behind this guide can easily be adapted to other domains and languages. 

\textbf{(b) \textsc{Athaar}, a corpus of 3,893 lines} manually
    transcribed at the line-level following this guideline.  As comparing witnesses is the foundation of textual criticism,  \textsc{Athaar} aligns multiple witnesses of the same text so that their divergences can be studied.
We report inter-annotators agreement  and analyze main transcription  errors.  We will make the corpus available upon acceptance. 
    
\textbf{(c) An evaluation of state-of-the-art
    vision-language HTR models.} We finally design a set of experiments to study model transfer across treatises, chapters and scripts. Our findings  highlight the limitations of current systems when faced with the specificities of ancient scientific texts. 







\section{Background}

\label{sec:terminology}

In  the remainder of this paper, we use the following terms, defined in textual philology
and digital humanities scholarship as follows. 

\paragraph{Treatise ($\mathrm{Tr}$).}
A written work devoted to a single subject, often organized into \textbf{books} (\textit{maqāla}), each subdivided into \textbf{chapters}.
Following the critical edition tradition \citep{loize2021}, we denote a
book-chapter pair by $\mathrm{B.ch}$ (e.g.,~X.6 for Book~X,
Chapter~6). 

\paragraph{Manuscript ($\mathrm{Ms}$).}
It refers to a handwritten document containing a whole treatise or only a part of it, as distinct from a printed book. In codicology,
it designates a physical object (a codex, scroll, or a bound volume) carrying a text written by hand \citep{deroche_islamic_2006}.
In the field of 
HTR, a manuscript is also the
primary object of digitization: high-resolution images of its pages
constitute the input to recognition systems.

\paragraph{Witness ($w$).}
Any source that preserves a state of a given
text, typically a manuscript copy, but also a printed edition, a
translation, or a quotation \citep{roelli_handbook_2020}. In this paper,
a witness is the copy of one chapter in one manuscript, so a single
manuscript can provide several witnesses (e.g., Ms01 provides X.6 and
I.4). Witnesses of the same text diverge because \textbf{scribes}, the
individuals who copied a manuscript by hand, as opposed to the text's
author, introduced errors, omissions, or deliberate changes.


\paragraph{Script ($\mathrm{scr}$).}
In paleography, \textit{script} denotes a specific calligraphic style, distinct from the notion of a writing system (e.g.\ the Arabic
script as a character repertoire). For Arabic manuscripts, the principal
book scripts include \textit{Naskh}, a clear rounded hand that became the
dominant bookhand from the tenth century onward; \textit{Nasta'liq}, a
style developed in Iran characterized by a strongly diagonal baseline and
fluid letter connections; and \textit{Maghrebi}, a regional variant
prevalent in North Africa and al-Andalus \citep{deroche_islamic_2006}. 

Figure~\ref{fig:witnesses} in Appendix \ref{sec:multi-witness variation}
illustrates the  same passage of Chapter~X.6  from the Qānūn treatise   across four manuscripts and two scripts.

\newcolumntype{Y}{>{\raggedright\arraybackslash}X}
\begin{table*}[h!]
    \centering
    \scriptsize
    \setlength{\tabcolsep}{2pt}
    \renewcommand{\arraystretch}{1.1}
    \begin{tabularx}{\textwidth}{>{\raggedright\arraybackslash}p{1.75cm} Y Y Y Y Y Y c}
    \toprule
    \textbf{Name} & \textbf{Domain} & \textbf{Date / period} & \textbf{Volume} & \textbf{Numeral content} & \textbf{Script variation} & \textbf{Several witnesses of the same text} & \textbf{Guide / agreement} \\
    \midrule
    \textbf{Muharaf}\newline\citep{saeed_muharaf_2024} &
    Literature &
    19th--21st c. &
    1,644 images; 36,311 lines &
    n/r &
    No: Only Ruq'ah &
    No &
    \textcolor{red}{\ding{55}} / \textcolor{red}{\ding{55}} \\
    \cmidrule[0.25pt]{1-8}
    \textbf{Agapet}\newline\citep{ibrahim_agapet_2024} &
    Religious &
    13th, 14th, 17th c. &
    831 pages &
    n/r &
    Yes, n/r &
    n/r &
    \textcolor{red}{\ding{55}} / \textcolor{red}{\ding{55}} \\
    \cmidrule[0.25pt]{1-8}
    \textbf{Historical Arabic HTR}\newline\citep{NAJAM2024110813} &
    Religious &
    n/r &
    40 pages; 640 lines &
    n/r &
    n/r &
    No &
    \textcolor{red}{\ding{55}} / \textcolor{red}{\ding{55}} \\
    \cmidrule[0.25pt]{1-8}
    \textbf{RASM}\newline\citep{rasm_2018} &
    Scientific &
    10th--19th c. &
    120 pages; 2,164--2,613 lines &
    Yes: n/r &
    n/r &
    n/r &
    \textcolor{red}{\ding{55}} / \textcolor{red}{\ding{55}} \\
    \cmidrule[0.25pt]{1-8}
    \textbf{RASAM}\newline\citep{vidalgorene:halshs-03430697} &
    Literature &
    17th--19th c. &
    300 images; 7,540 lines (v1) &
    Yes: fractions, tables; n/r &
    No: Only Maghrebi &
    No &
    \textcolor{red}{\ding{55}}$^{\dagger}$ / \textcolor{red}{\ding{55}} \\
    \cmidrule[0.25pt]{1-8}
    \textbf{Iskandar}\newline\citep{iskandar_2024} &
    Literature &
    17th--19th c. &
    297 pages &
    n/r &
    Yes, n/r &
    Yes: 5 copies of one work but not same content &
    \textcolor{red}{\ding{55}} / \textcolor{red}{\ding{55}} \\
    \midrule
    \textbf{\textsc{Athaar}}\newline(ours) &
    \textbf{Scientific: astronomy and mathematics} &
    \textbf{11th--19th c.} &
    \textbf{169 pages}, \textbf{3,893 lines} &
    Yes; \textbf{2,618 digits}, transcribed &
    Yes: \textbf{Naskh, Nasta'liq, Maghrebi} &
    Yes: \textbf{13 witnesses of chapter X.6} &
    \textcolor{green!60!black}{\ding{51}} / \textcolor{green!60!black}{\ding{51}} \\
    \bottomrule
    \end{tabularx}
    \caption{Arabic historical HTR datasets. n/r: not reported in the sources checked.
    The last column reports whether a an annotation guideline exists and whether
    inter-annotators agreement is measured (\textcolor{green!60!black}{\ding{51}}~yes,
    \textcolor{red}{\ding{55}}~no).
    $^{\dagger}$: RASAM documents general transcription choices but does not
    release a formal annotation guide.  }
    \label{tab:datasets_comparison}
\end{table*}

\section{Related Work} \label{sec:related-datasets}
This section focuses on Arabic resources and models for HTR. The reader can refer to \citet{sommerschield_machine_2023} and \citet{Garrido2026} for a comprehensive overview in the field. 
\paragraph{Existing 
Datasets.} Table~\ref{tab:datasets_comparison} surveys the Arabic historical HTR
datasets most relevant to our work. Beyond the domain and multi-witness
gaps noted in the Introduction, this table reveals two
further limitations. First, no dataset documents how numerals are handled during transcription: whether numeral strings are transcribed, tagged, or silently omitted. 
Although RASM \citep{rasm_2018} is the only one targeting
scientific manuscripts, no documentation describes its
transcription conventions, the scientific domains covered, or the
distribution of content across the 120 pages, making it impossible to
assess its suitability for training or evaluating recognition systems on
specific scientific text types.
Second, none reports agreement between
annotators, making it impossible to assess ground-truth quality or
compare results across datasets on equal terms.
\textsc{Athaar} is smaller than Muharaf or RASAM in total line count,
but it is designed for a different purpose: to enable controlled
evaluation of transfer across scribes, chapters, and treatises.
rather than
to maximize scale.

\paragraph{Dataset quality.} 
\label{sec:annotation-quality}
Measuring dataset quality in terms of inter-annotators agreement is not common in the field. Among the few studies, we cite 
\citet{tarride2023crowd} who performed 
two independent transcriptions on 37\% of their dataset 
composed of French
handwritten official meeting records and used the character error rate (CER) between them to select lines for their testsets. \citet{aresoliveira2018venetian}
compared crowd contributors with archivists on 18th-century Venetian
records and measured CER between the annotators which span from 10.5 to 13.3\% depending
on the normalization step. Beyond direct measurement, reports on ground-truth
quality are mostly production targets: the IMPACT and ENP datasets
specified 99.95\% accuracy for commissioned ground truth
\citep{papadopoulos2013impact, clausner2015enp}. \citet{neudecker2021survey}
further show that error rates depend on the evaluation tool and its
Unicode normalization, making cross-dataset comparisons unreliable without
a stated normalization protocol.

\paragraph{Arabic HTR Models.}
\label{sec:related-models}
Recent work 
falls into two groups. The first
covers printed and modern handwritten Arabic. Qalam
\citep{bhatia_qalam_2024} is a multimodal model fine-tuned on printed
benchmarks and synthetic data generated from modern Arabic books.
Qari-OCR-2B~\citep{wasfy_qari-ocr_2025} follows a similar approach, fine-tuning Qwen-2-VL-2B using a synthetic pipeline that renders text from modern news and classical Islamic corpora in twelve Arabic fonts. 
The second group targets handwritten and historical text. HATFormer
\citep{chan_hatformer_2025} fine-tunes a TrOCR-based architecture on
synthetic line images and Muharaf~\citep{saeed_muharaf_2024}, a corpus of 19th--21st century documents. CHURRO-3B \citep{semnani_churro_2025} is a fine-tuned Qwen2.5-VL-3B-Instruct on historical manuscripts across multiple languages. Arabic training data comes from  
existing datasets including Muharaf, RASAM and RASM for a total of 2,367 pages. None of these datasets covers scientific manuscripts, with specificities
such as script variation and multiple witnesses of the same text.

As far as we know, no existing model has been
evaluated on medieval Arabic
scientific manuscripts, which motivates both \textsc{Athaar} and the
experiments in Section~\ref{sec:experiments}.

\section{The \textsc{Athaar} 
Dataset}
\subsection{Sources}
\label{sec:sources}

Our dataset is composed of a set of scientific manuscripts sampled from different authors. Our main source is 
\textit{al-Qānūn al-Mas‘ūdī}, an
astronomical treatise composed by al-Bīrūnī around 1035 which  
 is organized into eleven books, each divided into chapters. Four of the manuscripts
of the Qānūn manuscripts are available
online, while the remaining copies were acquired 
from the holding libraries as part of the preparation of a critical edition of the Qānūn ~\cite{loize2021}. 
In addition,  the corpus also includes other scientific treatises from the same period covering astronomy and mathematics. To mitigate the risk of training-data contamination, we prioritized the collection of manuscript witnesses not publicly accessible online. 

\paragraph{Corpus structure.}
\textsc{Athaar} is organized into three tiers that
move progressively away from the anchor text, as shown in  Table~\ref{tab:manuscripts_summary}: \vspace{0.2cm}

\begin{table}[t]
\centering
\scriptsize
\setlength{\tabcolsep}{2pt}
\renewcommand{\arraystretch}{1.02}
\begin{tabularx}{\columnwidth}{>{\raggedright\arraybackslash}X
    >{\raggedright\arraybackslash}p{1.1cm}
    r
    >{\raggedright\arraybackslash}p{1.0cm}
    r r r}
\toprule
\textbf{Manuscript} & \textbf{Date} & \textbf{\#ch} & \textbf{Script} &
\textbf{\#Pg/Ln} & \textbf{\%D.} & \textbf{Dg.} \\
\midrule
\multicolumn{7}{l}{\textit{Tiers 1--2: al-Qānūn}} \\
Ms01 (Berlin, 1613)    & 12th c.  & 2 & Naskh     & 7/224   & 2.58  & 198 \\
Ms02 (Berlin, 275)     & 19th c.  & 2 & Nasta'liq & 11/290  & 0.08  & 200 \\
Ms04 (Cairo, 866)      & 13th c.  & 2 & Naskh     & 8/217   & 20.63 & 195 \\
Ms05 (Hyderabad, 374)  & 17th c.  & 1 & Nasta'liq & 8/156   & 0.00  & 190 \\
Ms06 (Istanbul, 87)    & 11--12th & 2 & Naskh     & 14/339  & 12.88 & 202 \\
Ms07 (Konya, 6625)     & 18th c.  & 1 & Naskh     & 8/238   & 0.01  & 173 \\
Ms08 (Istanbul, 1498)  & 12th c.  & 2 & Nasta'liq & 12/231  & 0.18  & 198 \\
Ms09 (Istanbul, 2277)  & 11th c.  & 1 & Naskh     & 4/93    & 0.12  & 38  \\
Ms10 (Istanbul, 65)    & 18th c.  & 2 & Nasta'liq & 9/180   & 0.50  & 198 \\
Ms13 (London, Or-1997)     & 12th c.  & 2 & Naskh     & 8/231   & 1.44  & 225 \\
Ms14 (Mashhad, 5588)   & 17th c.  & 1 & Naskh     & 6/134   & 2.06  & 185 \\
Ms15 (Mashhad, 12292)  & 13th c.  & 1 & Naskh     & 9/185   & 0.63  & 192 \\
Ms16 (Oxford, Or-516)    & 11th c.  & 1 & Naskh     & 3/75    & 2.20  & 23  \\
Ms17 (Paris, Ar. 6840)   & 12th c.  & 2 & Naskh     & 6/212   & 0.01  & 201 \\
Ms18 (Rampur, 3700)    & 17th c.  & 1 & Naskh     & 6/164   & 1.25  & 198 \\
\cmidrule{1-7}
\textbf{Sub. T1--T2}  &          & \textbf{23} & &
\textbf{119/2,969} & \textbf{3.75} & \textbf{2,616} \\
\midrule
\multicolumn{7}{l}{\textit{Tier 3: Other scientific treatises (BnF)}} \\
Ms (Paris, Ar 2494)  & 12th c. & 1 & Naskh    & 6/110  & 1.41  & 0 \\
Ms (Paris, Ar 2458)  & 12th c. & 1 & Naskh    & 3/82   & 3.02  & 0 \\
Ms (Paris, Ar 2504)  & 17--18th& 1 & Naskh    & 6/124  & 0.04  & 0 \\
Ms (Paris, Ar 2546)  & 17th c. & 1 & Naskh    & 6/127  & 0.04  & 0 \\
Ms (Paris, Ar 2497)  & 17th c. & 1 & Naskh    & 4/82   & 1.12  & 0 \\
Ms (Paris, Ar 2509)  & 14th c. & 1 & Naskh    & 4/60   & 15.04 & 0 \\
Ms (Paris, Ar 2500)  & 15th c. & 1 & Naskh    & 5/68   & 35.43 & 2 \\
Ms (Paris, Ar 2484)  & 13th c. & 1 & Naskh    & 3/72   & 0.00  & 0 \\
Ms (Paris, Ar 4821)  & 12th c. & 1 & Maghrebi & 13/199 & 1.99  & 0 \\
\cmidrule{1-7}
\textbf{Sub. T3}      &           & \textbf{9}  & &
\textbf{50/924}   & \textbf{6.24} & \textbf{2}    \\
\midrule
\textbf{Total}        &           & \textbf{32} & &
\textbf{169/3,893} & \textbf{4.19} & \textbf{2,618} \\
\bottomrule
\end{tabularx}
\caption{\textsc{Athaar} manuscripts
identified as \textit{Ms\,(Location, Shelfmark)}.
\#ch, Pg, Ln: number of chapters, pages and lines;
\%D.: diacritic marks as a share of non-space characters;
Dg.: transcribed digits.
}
\label{tab:manuscripts_summary}
\end{table}

\noindent \textbf{[Tier 1: same chapter, different scribes.]} The anchor is chapter X.6. We
  chose it because it continues the work of \citet{loize2021}, whose
  edition serves as our reference, and because it supports one of the project's aims of making a critical edition of \textit{al-Qānūn al-Mas‘ūdī}.
  Chapter X.6 is dedicated to the quantifications of celestial distances and volumes.
  To exhaustively cover this topic, al-Bīrūnī relays on Greek, Sanskrit and Arabic sources. Its 13 witnesses come from 13 manuscripts spanning a period from the 11th to the 19th century and written in
  Naskh and Nasta'liq.\footnote{In Table \ref{tab:manuscripts_summary},  Ms03 was excluded because 57\% of its lines contain illegible text, see Figure~\ref{fig:ms03} in the Appendix. Ms11 and Ms12 are not part of Tier1 as they omit Chapter X.6.}

\noindent \textbf{[Tier 2: same treatise, different chapters.]} Ten further chapters come from
  ten manuscripts, eight of which also provide a witness of X.6. For these
  eight, the manuscript is held fixed while the chapter changes. The chapters
  were chosen to vary the script and the type of content: some contain
  numerals or diacritics and others none. 

\noindent \textbf{[Tier 3: different treatises.]} To test transfer beyond the Qānūn, we added
  9 witnesses of other scientific treatises held by the Biblioth\`{e}que nationale de France (BnF) accessed via its digital library Gallica,\footnote{\url{gallica.bnf.fr}} written in the same period and belonging to different authors such as Ibn Sīnā, al-Farghānī, Ibn al-Haytham, al-T\={u}s\={\i}, al-Jaghm\={\i}n\={\i}, Ab\={u} al-Waf\={a} and al-B\={\i}r\={u}n\={\i}.

\paragraph{Corpus Challenges.} \label{challenge}
Besides script and scribe variations, transcribing this dataset presents several technical challenges, among which: (i) \textit{Layout}. The original manuscripts contain marginal notes, astronomical tables, and diagrams. 
(ii) 
    \textit{Paleography}. The script of the corpus differs radically from that of modern Arabic 
    and printed Arabic 
    with a frequent use of elongation (\textit{kashida} / \textit{tatweel}), 
    complex ligatures, and variation in the representation of numerals. 
    (iii)  \textit{Condition of the manuscripts}. Physical damage to the materials (water stains, faded ink, tears) introduces significant visual noise that the models must learn to filter out. Figures~\ref{fig:modern_print}, \ref{fig:modern_hand}, \ref{fig:historical_ms}  from Appendix \ref{sec:script_examples} illustrate these challenges.

\subsection{Dataset Statistics} \label{sec:stats}

Table~\ref{tab:manuscripts_summary} reports the statistics of the manuscripts, one row per chapter. Tier~1 contains 13 witnesses of chapter X.6, each from a different manuscript: 87 pages, 2,165 lines, 35,610 words and 152,065 characters--about 2,700 words per witness. Tier~2 contains 10 witnesses of other chapters from 10 manuscripts, eight of which also appear in Tier~1: 32 pages, 804 lines, 12,168 words and 54,120 characters. Tier~3 contains 9 witnesses of other scientific treatises, each from a different  manuscript: 50 pages, 924 lines, 10,377 words and 44,295 characters. In total, \textsc{Athaar} comprises 24 manuscripts, 32 witnesses, 169 pages, 3,893 lines, 58,155 words and 250,480 non-space characters. 

\subsection{Dataset Preparation}
Prior to the  annotation process, the digital copies of each manuscript underwent
five pre-processing steps:
(1)~rotation correction to align pages;
(2)~conversion to JPG format for compatibility with the annotation
platform (see below);
(3)~page splitting, separating double-page scans into individual folios;
(4)~image cropping to a consistent resolution within each manuscript; and
(5)~scheduling of annotation batches by manuscript difficulty, grouping
witnesses with high diacritization rates, dense digit strings, or
significant physical degradation (ink fading, bleed-through) into
dedicated weekly sessions to allow annotators to calibrate their
judgement progressively.
Annotation is performed at the line level, following the PAGE-XML standard
\citep{pletschacher_page_2010} promoted by the PRImA
Research Lab, which is the de facto granularity for HTR evaluation.

\subsection{Annotation Procedure}
\label{sec:guidelines}

\paragraph{Guidelines.} Annotators followed a written guide, provided in Appendix~\ref{sec:excerpt_guide} and in the supplementary material. 
It was written for Arabic scientific manuscripts, drawing on the
principles of diplomatic editing \citep{west_textual_1973}: faithful
reproduction of what the scribe wrote without correcting, modernizing the spelling or
interpreting the content, preservation of domain-specific
notation (numerals, geometrical objects), and
explicit rules for every element that deviates from running prose. The
critical edition of \citet{loize2021} served as the textual reference,
supplying the authoritative reading against which each transcribed line
is aligned. The same guide was applied uniformly across all 
tiers.


Annotators are asked to transcribe each line directly from the image 
, in
reading order, from top to bottom and right to left. To reduce complexity due to layout (cf. Section \ref{challenge}),  only the titles and main text are transcribed.
Running headers, page footers, page and folio numbers, independent marginal notes (additions, glosses, comments), tables, diagrams, ornaments and library stamps are excluded, and regions detected automatically in these categories are deleted. One exception applies: when a word begun in the main text is completed in the margin, the two fragments are joined in the transcription of that line, and the line polygon is extended to include the margin fragment.

Regarding diacritics and vowel signs, they  are transcribed when the scribe wrote them and are never added elsewhere. Finally, 
scientific manuscripts contain many numerical values, often overlined. A numeral the annotator can read is transcribed. Otherwise the annotator inserts a tag naming its numeral system (e.g., \texttt{[indian\_numbers]}), or an unknown-system tag (e.g.,\texttt{[unknown\_numbers]}) when the system cannot be identified. Other overlined elements that denote geometrical objects receive their own tag (e.g.,\texttt{[geometrical\_object]}). Damaged or illegible text is replaced by a tag (e.g.,\texttt{[unreadable]}), either within the line or for the whole line (see details in Appendix \ref{t-guidelines}). 


\paragraph{Tool.} We used Calfa Vision\footnote{\url{https://vision.calfa.fr}} \citep{vidalgorene:hal-03971691}, a web platform for assisted transcription of handwritten and printed documents   for three reasons: it provides a generic layout model for historical Arabic manuscripts, whose output the annotator corrects instead of drawing lines from scratch; it was already used to build Arabic manuscript datasets such as RASAM \citep{vidalgorene:halshs-03430697} and it exports PAGE-XML, the format of most datasets in Table~\ref{tab:datasets_comparison}. The transcription rules do not depend on the tool: any editor that exports PAGE-XML can apply them. The tool is illustrated in  Figure~\ref{fig:calfa_platform}, Appendix~\ref{sec:manual_annotation}.

\paragraph{Annotation phases.}
Annotation was carried out by two authors of this paper which are native Arabic speakers with expertise in medieval Arabic manuscripts, under the supervision of an expert in Arabic philology, following a three steps procedure.  In the first  \textit{training phase}, each annotator independently transcribed 8 pages from Tier 1 (Ms04, Ms05, Ms07, Ms10, Ms14 and Ms15)
using Calfa Vision and following the annotation guide. The supervisor reviewed the results, provided corrections and feedback, and confirmed that both annotators had reached a sufficient understanding   of the
guide before moving to the next step. 

In the \textit{adjudication phase}, both annotators independently transcribed the same 4 pages - one per manuscript, across Ms01, Ms02, Ms13 and Ms17 (127 lines) corresponding to 4\% of the total lines of Tiers~1 +~2. 
These pages have been chosen  as they contain full text without any additional content (tables, diagrams), covering more text and making the results comparable. The agreement between the two transcriptions was then computed and is reported in Section~\ref{sec:quality}. The supervisor verified that the agreement metrics were satisfactory before authorizing the full annotation.

In the \textit{full annotation phase}, each page image was first checked for sharpness, completeness and orientation. Text regions
and line baselines were then detected automatically by Calfa Vision's
built-in layout analysis model. 
The annotator removed out-of-scope regions, merged or split
regions, and corrected missing, merged or misplaced baselines; line polygons were then computed from the corrected baselines and checked visually. Finally, the annotator transcribed   at the line level.
Each manuscript was exported as PAGE-XML, which stores the baseline, the polygon and the Unicode transcription of every line, together with the exact page images so that coordinates and images remain aligned.
Annotation proceeded in weekly batches, and the supervisor validated each batch before the next one was assigned.

\section{Results of the Annotation Campaign}

\subsection{Qualitative Results} 
\label{sec:quality}
Following \citet{tarride2023crowd} and \citet{aresoliveira2018venetian}, inter-annotators agreements are computed in terms of  
 the Character Error Rate (CER) and Word
Error Rate (WER), which are  based on the Levenshtein distance between the
transcription and the reference text (lower is better). Before computing these metrics, data are first normalized by collapsing
redundant blank characters and removing special characters. The rates are then
computed by concatenating the lines of each page in reading order and
aligning the two transcriptions. Table \ref{tab:quality} reports our results.


A CER of 2.6\% between two careful
transcriptions of the same pages indicates that the images are largely
unambiguous and that the two annotators read them consistently. This
figure serves as a measure of the quality of \textsc{Athaar}'s
annotations rather than a domain standard: to our knowledge, no prior
Arabic HTR dataset reports annotator agreement, so no established
threshold exists for comparison \citep{tarride2023crowd, aresoliveira2018venetian}.
It is best understood as a reference point against which future recognition results on this corpus can be interpreted. 

Disagreement
nonetheless varies across witnesses: CER ranges from 1.6\% on Ms01 to
4.5\% on Ms13, although Ms01, Ms13 and Ms17 are all written in Naskh,
while the Nasta'liq witness Ms02 stays at 2.4\%. A shared script label
therefore does not imply a similar reading difficulty: the individual
scribe and the condition of the page also matter. This is precisely why a multi-witness corpus is more informative than a single-copy dataset: two manuscripts labelled Naskh can present different levels of difficulty for both human readers and recognition models, and a system trained or evaluated on only one of them will not generalize to the other. Appendix~\ref{sec:annotation_challenges} further details main cases of disagreements between annotators. 

\begin{table}[t]
\centering
\small
\setlength{\tabcolsep}{4pt}
\renewcommand{\arraystretch}{1.1}
\begin{tabular}{lrcc}
\toprule
\textbf{Manuscript} & \textbf{Lines} & \textbf{CER (\%)} $\downarrow$ & \textbf{WER (\%)} $\downarrow$ \\
\midrule
Ms01 & 33 & 1.6 & 7.1 \\
Ms02 & 27 & 2.4 & 9.2 \\
Ms13 & 31 & 4.5 & 20.5 \\
Ms17 & 36 & 2.2 & 12.2 \\
\midrule
\textbf{Pooled} & \textbf{127} & \textbf{2.6} & \textbf{11.7} \\
\bottomrule
\end{tabular}
\caption{Inter-annotators agreement on the first page of four witnesses
of chapter X.6. 
}
\label{tab:quality}
\end{table}

\subsection{Quantitative Results}
\label{sec:characteristics}
The final dataset has the following characteristics (see Table~\ref{tab:manuscripts_stats} Appendix~\ref{sec:corpus_stats}). 

\paragraph{Scribes and scripts.} The 15 manuscripts of the Qānūn were copied by different scribes between the 11th and the 19th century:
eleven in Naskh and four in Nasta'liq (Ms02, Ms05, Ms08, Ms10), which account for 29\% of the lines. Since the 13 witnesses of X.6 carry nearly the same text (except some variants), their differences come from the script, not from the content, and even within one script the scribes differ visibly (Figure~\ref{fig:witnesses}).

\paragraph{Length and density.} The same chapter occupies 4 to 9 pages and 134
to 238 lines depending on the witness, from 13 to 22 words per line.

\paragraph{Diacritics marks.} They make up 4.19\% of the non-space characters. They are unevenly distributed: four manuscripts (Ms04, Ms06 and, in Tier~3, Ms(Paris, Ar 2500) and Ms(Paris, Ar 2509)) hold 83\% of all marks, which make up 9.9 to 35.4\% of their characters, whereas 13 of the 32 witnesses stay below 0.3\%. Vocalized and unvocalized text thus coexist in the corpus, within one treatise and across treatises, which makes it possible to test the transfer from one to the other.

\paragraph{Numerals.} The texts contain 688 numeral strings (about 12 per 1,000 words, 2,618 digits in total), and all but two are transcribed. Numerals concentrate in X.6: each of its 13 witnesses holds 48 to 56 strings, whereas seven of the ten Tier~2 witnesses and eight of the nine Tier~3 witnesses contain none. Since they are transcribed with Eastern Arabic-Indic digits, numerals can be scored separately from the running text.

\paragraph{Condition.} 
When the text remained legible to the annotator, it is transcribed; the
\texttt{[unreadable]} tag marks only the passages in which the word could not be
recovered, so the tag counts give a lower bound of the damage. Only 1\% of the
lines (30 of 2,969) carry a tag, and three manuscripts (Ms05, Ms08, Ms17) hold 27
of the 33 \texttt{[unreadable]} tags. 


\section{Experiments}
\label{sec:experiments}


\subsection{Experimental Protocol}
Our aim is to evaluate transfer capabilities of state-of-the art VLMs models across scripts, chapters and treatise. Let  $w$ be a  witnessis defined by its manuscript
$\mathrm{ms}(w)$,  chapter $\mathrm{ch}(w)$ and  script
$\mathrm{scr}(w)$; $L(w)$ denotes its lines. The tiers are
$T_1 = \{w : \mathrm{ch}(w) = \text{X.6}\}$, $T_2$ the witnesses of other
chapters of the \textit{Q\=an\=un}, and $T_3$ the witnesses of other
treatises. Following standard few-shot terminology \citep{vinyals2017},
the lines of each witness are split once, with a fixed seed, into a
support part $L^S(w)$ (30\% of its lines, drawn randomly) and a query
part $L^Q(w) = L(w) \setminus L^S(w)$. Each experiment is run at $k \in \{0, 3\}$ (see Section~\ref{sec:k_ablation}, Appendix \ref{sec:experiment_extra}): the model receives $k$ exemplars drawn
from a support bank $\mathcal{B}$ and is evaluated on all lines of
$L^Q(w)$ for each query witness $w$. Support and query lines never come from the same witness, so they are
always disjoint (see Table~\ref{tab:experiment_sizes} for testset split).

We design three transfer scenarios that leverage
the tiered structure of \textsc{Athaar}, see Section~\ref{sec:splits}.

\begin{enumerate}
    \item \textbf{Cross-script (\textsc{C-Scr}):} We hold the chapter constant (X.6) while evaluating across witnesses from $T_1$ written in a different
    script among Naskh ($T_1^{N}$) and
  Nasta'liq ($T_1^{NT}$). Models are thus instructed with a support $\mathcal{B} = L^S(w)$, where  $w \in T_1^{N}$ and tested on   $w' \in T_1^{NT}$, and vice versa. Exemplars are drawn once per query witness.
  This setting aims to isolate the effect of script change on recognition quality.  

    \item \textbf{Cross-chapter (\textsc{C-Ch}):} We use manuscripts that appear
    in both Tier~1 and Tier~2, holding the scribe fixed while the
    chapter changes, isolating whether models
    transfer across distinct content. Thus models are instructed with examples $\mathcal{B} = L^S(w)$ such as $w \in T_1$ and tested on $w' \in T_2$ where $\mathrm{ms}(w) = \mathrm{ms}(w')$ and  $\mathrm{ch}(w) \neq \mathrm{ch}(w')$.
  Again, exemplars are drawn once per
  query witness.\footnote{Experiments were made with support  drawn from $T_1$ and query from $T_2$ and not vice versa,
because $T_1$ is the anchor set: its 13 witnesses of chapter~X.6 form
the richest and most uniformly annotated pool in the corpus, making it
the natural source of exemplars for  transfer.}

    \item \textbf{Cross-treatise (\textsc{C-Tr}):} We evaluate out-of-domain
    generalization to entirely unseen scientific treatises, providing
    support exemplars from \textit{al-Q\=an\=un} and evaluating on
    Tier~3, i.e., support $\mathcal{B} = L^S(w)$ contains examples from  $w \in T_1 \cup T_2$ and the testset manuscripts $w' \in T_3$.
\end{enumerate}

\subsection{Models and Settings}
\label{sec:models_settings}
We evaluated several models from three families (general-purpose multimodal models, Arabic-oriented VLMs, and OCR-specialized VLMs) but only reports the best preforming ones: (1) \textbf{Gemini~3~Flash}
\citep{gemini3flash} a general-purpose VLM optimized for
speed and long-context processing. (2) \textbf{CHURRO-3B} \citep{semnani_churro_2025} an open-weights VLM for historical document transcription in 46 languages, including Arabic.
(3) \textbf{Qari-OCR-2B} \citep{wasfy_qari-ocr_2025} a VLM fine-tuned
specifically for Arabic text recognition. 

Models receive as input line crops, as extracted from the PAGE-XML
annotations using the polygon coordinates of each \texttt{TextLine} element.
They are prompted in English, 
preliminary experiments with Arabic prompts were not conclusive.

We run each model in both zero-shot and few-shot settings 
(see Appendix \ref{sec:k_ablation} for an ablation study).  
We report the average CER and WER across 3 runs after normalizing models outputs and the reference annotation (ground truth) following the standard protocol in the field:~redundant blank characters are collapsed;
~special characters are removed;
~Numbers are normalized to their Arabic-Indic equivalents; and
~any characters outside the Arabic script are removed. An important pre-processing prior to evaluation was removing \textit{tatweel} (elongation). Indeed,  scribal elongation can trigger runaway
generation: in one \textsc{C-Ch} line, the scribe's extended \textit{tatweel} led Gemini to generate 4,037 characters instead of 11, a single error that alone raised its CER on \textsc{C-Ch} from 18.9\% to 74.3\%, illustrating
why normalization before scoring is necessary. Table~\ref{tab:main_results} reports our results. 

\subsection{Results}
\label{sec:results}

\begin{table}[t]
\centering
\small
\setlength{\tabcolsep}{3pt}
\renewcommand{\arraystretch}{1.1}
\begin{tabular}{llcccccc}
\toprule
& & \multicolumn{2}{c}{\textbf{Gemini}}
  & \multicolumn{2}{c}{\textbf{CHURRO}}
  & \multicolumn{2}{c}{\textbf{Qari-OCR}} \\
\cmidrule(lr){3-4}\cmidrule(lr){5-6}\cmidrule(lr){7-8}
\textbf{Exp.} & \textbf{$k$} & CER & WER & CER & WER & CER & WER \\
\midrule
\multirow{2}{*}{\textsc{C-Scr}(N)}
  & 0 & 17.2 & \textbf{62.1} & 76.5 & 124.5 & \textbf{88.6} & \textbf{135.5} \\
  & 3 & \textbf{16.3} & 62.3 & \textbf{70.0} & \textbf{115.6} & 154.5 & 198.9 \\
\addlinespace[2pt]
\multirow{2}{*}{\textsc{C-Scr}(NT)}
  & 0 & 20.0 & 55.4 & 87.3 & 143.4 & \textbf{93.9} & \textbf{157.6} \\
  & 3 & \textbf{18.5} & \textbf{53.8} & \textbf{79.9} & \textbf{134.9} & 158.6 & 219.5 \\ \hline
\addlinespace[2pt]
\multirow{2}{*}{\textsc{C-Ch}}
  & 0 & \textbf{18.9} & \textbf{61.8} & 75.3 & 127.5 & \textbf{78.7} & \textbf{127.7} \\
  & 3 & \textbf{18.9} & 63.2 & \textbf{59.3} & \textbf{109.2} & 163.1 & 194.0 \\
\addlinespace[2pt]
\multirow{2}{*}{\textsc{C-Tr}}
  & 0 & 21.7 & 64.3 & 120.3 & 153.8 & \textbf{100.7} & \textbf{164.9} \\
  & 3 & \textbf{20.4} & \textbf{61.1} & \textbf{63.9} & \textbf{112.5} & 252.9 & 260.8 \\
\bottomrule
\end{tabular}
\caption{CER and WER (\%). Best scores are in bold (lower is better) at zero-shot ($k=0$) and
few-shot ($k=3$). \textsc{C-Scr(a)}:  query witnesses are either in \textsc{a}=Naskh or \textsc{a}=Nasta'liq. 
Values above 100\% indicate outputs
longer than the reference.}
\label{tab:main_results}
\end{table}

\paragraph{Gemini transfers across all three settings.} Gemini~3~Flash
obtains the lowest error rates in every experiment at both $k=0$ and $k=3$, with a CER between
16.3\% and 21.7\%. Moving from witnesses of the same chapter (\textsc{C-Scr}) to
entirely different treatises (\textsc{C-Tr}) raises its CER by only 2 to 4 points,
indicating that its performance depends little on whether the text
belongs to the \textit{Q\=an\=un}. Its WER is three to four times its CER, as expected for Arabic: a single misread dot or letter invalidates the whole word.

\paragraph{Script direction.} Within \textsc{C-Scr}, the CER/WER asymmetry
holds at both $k$ values: the best scores of CER are lower on Naskh queries
(N:16.3--17.2\% vs.\ NT:18.5--20.0\%), while WER is lower on Nasta'liq
queries (NT:53.8--55.4\% vs.\ N:62.1--62.3\%). This pattern is consistent
but modest, and the two metrics point in opposite directions, so
neither script can be declared uniformly harder.

\paragraph{Specialized models do not transfer well.} CHURRO reaches a CER of 75.3--120.3\% at $k=0$ and improves to 59.3--87.3\% at $k=3$, still three to four times that of Gemini, and a WER above 100\% in every setting.
Qari-OCR behaves differently: it is bounded at $k=0$
(CER 78.7--100.7\%) but degrades sharply at $k=3$ (CER
154.5--252.9\%), suggesting that its few-shot scenario destabilizes generation rather than improving it.
These scores reflect a failure
to produce a bounded transcription rather than recognition accuracy.
Neither specialization for historical documents nor for Arabic script translates into usable performance on medieval scientific manuscripts.


\section{Conclusion}
This work introduced \textbf{\textsc{Athaar}}, the first manually
annotated dataset with a documented annotation guide for medieval Arabic
scientific manuscript recognition, and the only historical Arabic dataset
offering multi-witness transcriptions of the same text. We built a
multi-tiered corpus of 3,893 lines 
and evaluated state-of-the-art models 
in three
transfer scenarios: same chapter across different scribal hands,
same manuscript across different chapters, and chapters from different treatises
entirely.

Our results show that general-purpose VLMs such as Gemini~3~Flash
dominate the current state of the art but still lag significantly as
manuscript complexity increases - ligatures, tatweel - even when support images are
provided. Models fine-tuned on the HTR task, by contrast, struggle to
reliably transcribe lines without hallucinating or omitting words.

Looking ahead, two directions appear most promising: fine-tuning
open-weights VLMs on \textsc{Athaar} with training strategies that balance
performance across scripts and content types; and developing methods that
explicitly model calligraphic specificities that remain beyond the reach of current
models. Finally, beyond HTR, \textsc{Athaar} opens directions for two further tasks:
\textit{line segmentation}, supported by ground-truth polygon coordinates, 
and \textit{script identification}, leveraging multi-witness
contrasts between \textit{Naskh} and \textit{Nasta'liq}.
\section*{Limitations}
This study has several limitations. First, the corpus is limited to a authors from the same period, restricting generalization to the
broader landscape of medieval Arabic scientific manuscripts. Expanding
\textsc{Athaar} to other authors, disciplines, and manuscript traditions is a
natural next step. 

Furthermore, while we evaluate both zero-shot and
few-shot settings, no fine-tuning experiments are conducted; supervised
or semi-supervised approaches will need to be explored to fully capture
the potential of domain-adapted models. 

Finally, an in-depth study of prompting strategies (such as heuristic exemplar selection), and extension to other VLM families is left for
future work.

\section*{Ethical Considerations}
This study used existing manuscripts freely available in national libraries all over the world, as part of the critical edition project we rely on. Our collection is only composed of scientific content. 

As a small portion of \textsc{Athaar} is accessible online, we had a careful check of the sources of our data to prevent from training-data contamination.

Annotators are co-authors of this paper and have been granted as part of their doctoral studies in Arabic historical manuscript transcription. 

\section*{Acknowledgement}
This work was carried out as part of the \textit{Athaar} project, which receives government funding administered by the National Research Agency (ANR) under the
France 2030 program, grant number “ANR-22-EXES-0015,” and is co-funded by the
Occitanie Region and the European Regional Development Fund. We would also like to thank OcciData\footnote{\url{https://occidata.irit.fr}} for providing the computing resources that enabled us to conduct the experiments on their infrastructure.
\bibliography{custom}
\appendix

\section{\textsc{Athaar} corpus}
\label{sec:athaar_stats}

\subsection{Illustrative Script Examples}
\label{sec:script_examples}
\begin{figure}[h]
    \centering
    \begin{subfigure}[b]{0.48\linewidth}
        \includegraphics[width=\linewidth]{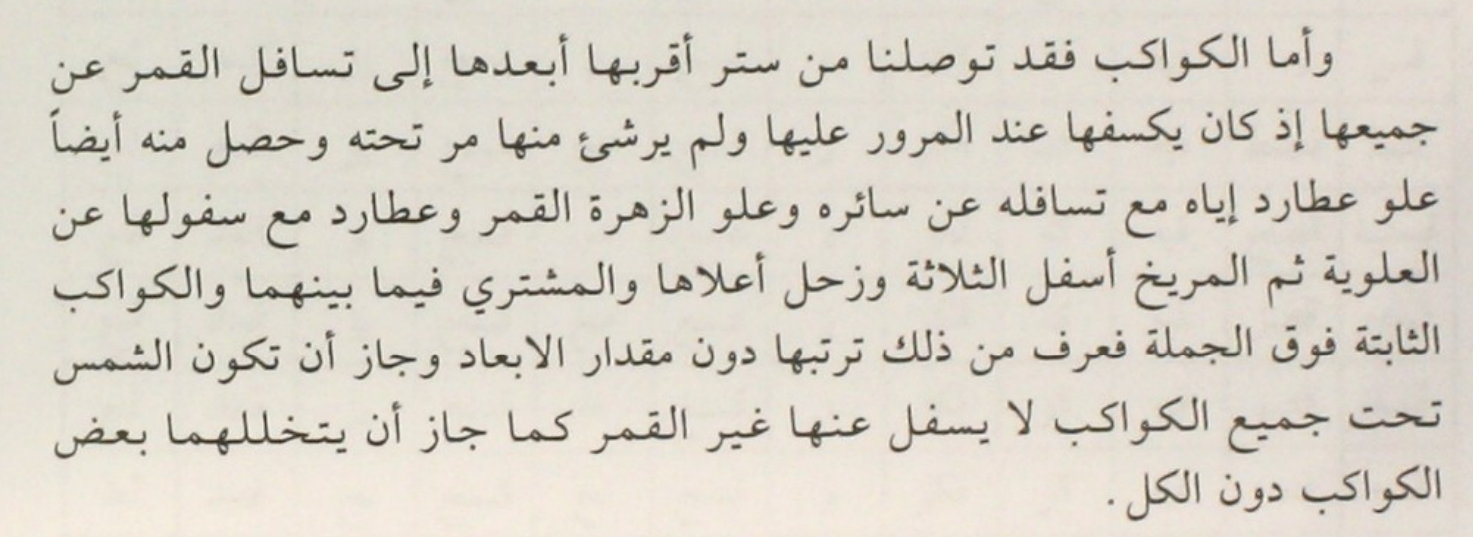} 
        \caption{Modern printed Arabic}
        \label{fig:modern_print}
    \end{subfigure}
    \hfill
    \begin{subfigure}[b]{0.45\linewidth}
        \includegraphics[width=\linewidth]{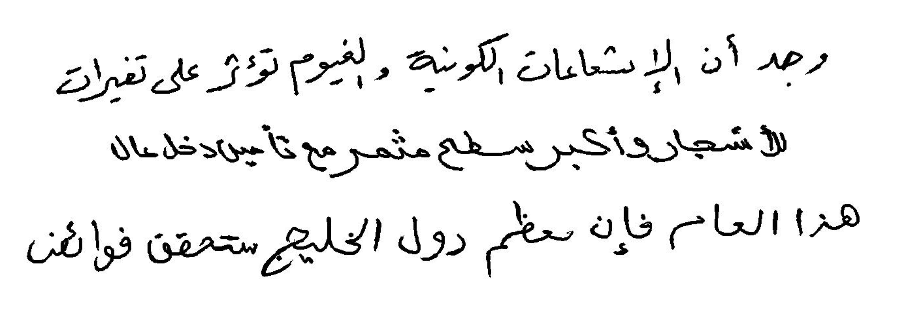}
        \caption{Modern Standard Arabic script \cite{mahmoud_khatt_2012}}
        \label{fig:modern_hand}
    \end{subfigure}
    \vspace{1cm} 
    \begin{subfigure}{\linewidth}
        \includegraphics[width=\linewidth]{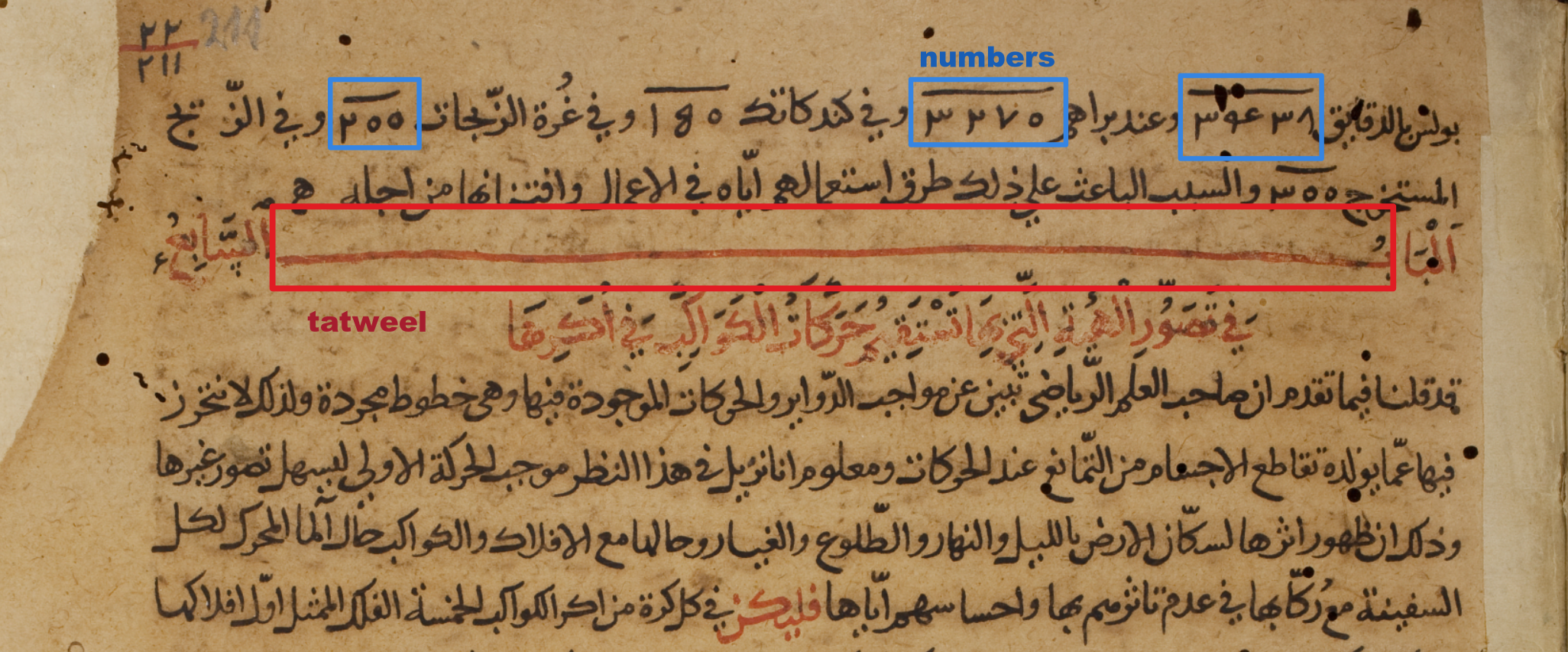}
        \caption{Excerpt from the corpus \textit{Qānūn} (Ms01).}
        \label{fig:historical_ms}
    \end{subfigure}
    \caption{Examples of difficulties found in manuscripts (cursive script, numerals, \textit{tatweel}).}
    \label{fig:script_comparison}
\end{figure}

Figure~\ref{fig:script_comparison} contrasts three writing contexts to illustrate the domain gap that motivates \textsc{Athaar}.
Figure~\ref{fig:modern_print} shows a modern arabic text: characters are uniform, spacing is regular, and
diacritics are placed consistently.
Figure~\ref{fig:modern_hand} shows modern handwritten Arabic \citep{mahmoud_khatt_2012}: letterforms vary between writers, but the script remains familiar to current OCR and HTR systems trained on modern data.
Figure~\ref{fig:historical_ms} shows an excerpt from Ms01 (12th~c., Berlin, Or-Quart-1613): the scribe uses prolonged \textit{tatweel}
strokes that extend words horizontally, mixes numeral strings with dense cursive text, and employs a calligraphic hand that has no direct equivalent in modern training corpora.
The visual distance between Figure~\ref{fig:modern_print} and
Figure~\ref{fig:historical_ms} explains why models trained or evaluated on modern Arabic fail to transfer to medieval scientific manuscripts.

\subsection{Multi-Witness Variation}
\label{sec:multi-witness variation}
\begin{figure}[t]
  \centering
  \begin{subfigure}[t]{0.48\textwidth}
    \includegraphics[width=\linewidth]{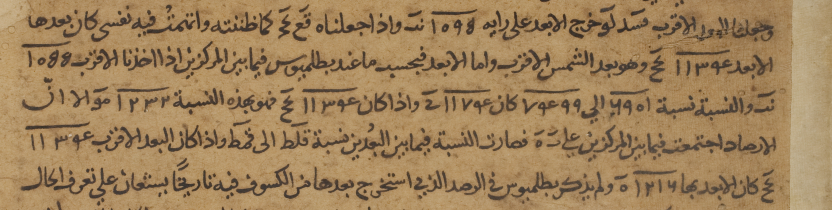}
    \caption{Ms01}
    \label{fig:ms01}
  \end{subfigure}
  \begin{subfigure}[t]{0.48\textwidth}
    \includegraphics[width=\linewidth]{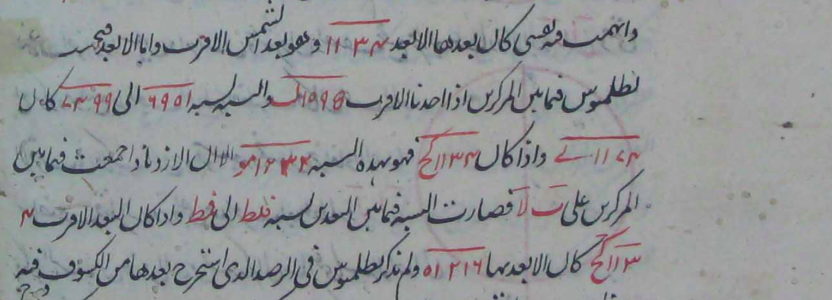}
    \caption{Ms05}
    \label{fig:ms05}
  \end{subfigure}
  \begin{subfigure}[t]{0.48\textwidth}
    \includegraphics[width=\linewidth]{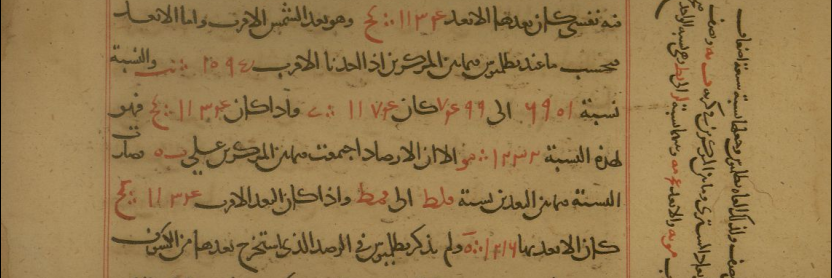}
    \caption{Ms15}
    \label{fig:ms15}
  \end{subfigure}
  \begin{subfigure}[t]{0.48\textwidth}
    \includegraphics[width=\linewidth]{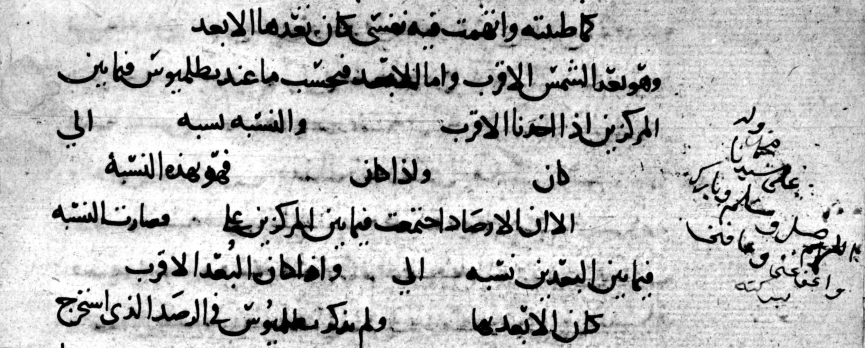}
    \caption{Ms03}
    \label{fig:ms03}
  \end{subfigure}
  \caption{The same passage of Chapter~X.6 across four manuscripts.
Ms03 was excluded from the corpus due to severe degradation.}
  \label{fig:witnesses}
\end{figure}

Figure~\ref{fig:witnesses} illustrates the variation in the same passage
of Chapter~X.6 across four manuscripts. Even within a single script
(Naskh: \ref{fig:ms01} and \ref{fig:ms15}), scribal hands differ
noticeably; the Nasta'liq hand (\ref{fig:ms05}) diverges further still.
Ms03 (\ref{fig:ms03}) was excluded from the corpus: bleed-through and
ink stains render part of the text unrecoverable. 

\subsection{Corpus Statistics}
\label{sec:corpus_stats}
Table~\ref{tab:manuscripts_stats} reports, for each manuscript, the number of pages, lines, words, and non-space characters, the diacritization rate (diacritized characters as a percentage of total non-space characters), numeral and digit counts, and the number of editorial markers (e.g.,~\texttt{[unreadable]}).
Manuscripts are grouped into two sections: (1) Tiers~1 and~2, drawn from \textit{al-Q\=an\=un al-Mas\textsuperscript{c}\=ud\={\i}}, with the number of annotated chapters per manuscript; (2) Tier~3, drawn from the BnF collection.

\begin{table*}[p]
    \centering
    \scriptsize
    \setlength{\tabcolsep}{1.8pt}
    \renewcommand{\arraystretch}{1.05}
    \begin{tabular}{>{\raggedright\arraybackslash}p{2.9cm} >{\raggedright\arraybackslash}p{1.8cm} c l l rrrrrrrrr}
    \toprule
    \multirow{2}{*}{\textbf{Manuscript}} & \multirow{2}{*}{\textbf{Date}} &
    \multirow{2}{*}{\textbf{\#ch.}} &
    \multirow{2}{*}{\textbf{Chapter}} & \multirow{2}{*}{\textbf{Script}} &
    \multicolumn{9}{c}{\textbf{Statistics}} \\
    \cmidrule(lr){6-14}
     & & & & &
    \textbf{Pages} & \textbf{Lines} & \textbf{Words} & \textbf{Chars} &
    \textbf{\shortstack[r]{Diacr.\\marks}} & \textbf{\shortstack[r]{\%\\diacr.}} &
    \textbf{\shortstack[r]{\hspace{5pt}Numeral\\strings}} & \textbf{Digits} & \textbf{Tags} \\
    \midrule
    \multicolumn{14}{l}{\textit{Tiers 1--2: \textit{al-Qānūn al-Mas‘ūdī}}} \\
    \addlinespace[2pt]
    \multirow{2}{2.9cm}{\raggedright Ms01\us Berlin\us Or-Quart-1613} & \multirow{2}{1.8cm}{\raggedright 12th c., 1166} & \multirow{2}{*}{2} & X.6 & \multirow{2}{*}{Naskh} & 5 & 158 & 3,114 & 13,461 & 316 & 2.35 & 51 & 198 & 3 \\
     & & & I.4 & & 2 & 66 & 1,298 & 6,079 & 189 & 3.11 & 0 & 0 & 0 \\
    \addlinespace[2pt]
    \cmidrule[0.25pt]{1-14}
    \multirow{2}{2.9cm}{\raggedright Ms02\us Berlin\us Or-Oct-275} & \multirow{2}{1.8cm}{\raggedright 19th c., 1834} & \multirow{2}{*}{2} & X.6 & \multirow{2}{*}{Nasta'liq} & 7 & 180 & 2,672 & 10,964 & 14 & 0.13 & 52 & 200 & 0 \\
     & & & XI.2 & & 4 & 110 & 1,519 & 6,881 & 1 & 0.01 & 0 & 0 & 0 \\
    \addlinespace[2pt]
    \cmidrule[0.25pt]{1-14}
    \multirow{2}{2.9cm}{\raggedright Ms04\us Cairo\us Talat-Miqat-DM-866} & \multirow{2}{1.8cm}{\raggedright 13th c., 1274} & \multirow{2}{*}{2} & X.6 & \multirow{2}{*}{Naskh} & 6 & 174 & 2,826 & 14,231 & 2,643 & 18.57 & 53 & 195 & 0 \\
     & & & IV.13 & & 2 & 43 & 742 & 4,208 & 1,161 & 27.59 & 0 & 0 & 0 \\
    \addlinespace[2pt]
    \cmidrule[0.25pt]{1-14}
    Ms05\us Hyderabad\us Asafiya-Riyada-374 & 17th c. & 1 & X.6 & Nasta'liq & 8 & 156 & 2,582 & 10,669 & 0 & 0.00 & 49 & 190 & 12 \\
    \addlinespace[2pt]
    \cmidrule[0.25pt]{1-14}
    \multirow{2}{2.9cm}{\raggedright Ms06\us Istanbul\us Askeri-Muze-87} & \multirow{2}{1.8cm}{\raggedright 11th--\allowbreak 12th c., $<$1137} & \multirow{2}{*}{2} & X.6 & \multirow{2}{*}{Naskh} & 8 & 191 & 2,523 & 11,586 & 1,151 & 9.93 & 51 & 202 & 0 \\
     & & & IV.4 & & 6 & 148 & 1,737 & 8,322 & 1,414 & 16.99 & 0 & 0 & 0 \\
    \addlinespace[2pt]
    \cmidrule[0.25pt]{1-14}
    Ms07\us Konya\us Yusufaga-42Yu6625 & 18th c. & 1 & X.6 & Naskh & 8 & 238 & 3,116 & 13,158 & 1 & 0.01 & 48 & 173 & 2 \\
    \addlinespace[2pt]
    \cmidrule[0.25pt]{1-14}
    \multirow{2}{2.9cm}{\raggedright Ms08\us Istanbul\us Suleymaniye-Jarullah-1498} & \multirow{2}{1.8cm}{\raggedright 12th c., 1136} & \multirow{2}{*}{2} & X.6 & \multirow{2}{*}{Nasta'liq} & 8 & 153 & 2,535 & 10,503 & 28 & 0.27 & 52 & 198 & 1 \\
     & & & V.1 & & 4 & 78 & 1,248 & 5,323 & 0 & 0.00 & 0 & 0 & 9 \\
    \addlinespace[11pt]
    \cmidrule[0.25pt]{1-14}
    Ms09\us Istanbul\us Beyazit-Veliyyuddin-2277 & 11th c., 1045 & 1 & V.7 & Naskh & 4 & 93 & 1,262 & 5,095 & 6 & 0.12 & 6 & 38 & 0 \\
    \addlinespace[2pt]
    \cmidrule[0.25pt]{1-14}
    \multirow{2}{2.9cm}{\raggedright Ms10\us Istanbul\us Kandilli-65} & \multirow{2}{1.8cm}{\raggedright 18th c., 1734} & \multirow{2}{*}{2} & X.6 & \multirow{2}{*}{Nasta'liq} & 7 & 142 & 2,695 & 11,189 & 65 & 0.58 & 55 & 198 & 0 \\
     & & & IX.6 & & 2 & 38 & 724 & 3,225 & 7 & 0.22 & 0 & 0 & 0 \\
    \addlinespace[2pt]
    \cmidrule[0.25pt]{1-14}
    \multirow{2}{2.9cm}{\raggedright Ms13\us London\us BL-Or-1997} & \multirow{2}{1.8cm}{\raggedright 12th c., 1174} & \multirow{2}{*}{2} & X.6 & \multirow{2}{*}{Naskh} & 5 & 146 & 2,521 & 10,554 & 114 & 1.08 & 51 & 198 & 0 \\
     & & & VI.9 & & 3 & 85 & 1,231 & 5,264 & 113 & 2.15 & 4 & 27 & 0 \\
    \addlinespace[2pt]
    \cmidrule[0.25pt]{1-14}
    Ms14\us Mashhad\us 5588 & 17th c., 1686 & 1 & X.6 & Naskh & 6 & 134 & 2,571 & 10,948 & 225 & 2.06 & 56 & 185 & 2 \\
    \addlinespace[2pt]
    \cmidrule[0.25pt]{1-14}
    Ms15\us Mashhad\us Radawi-12292 & 13th c. & 1 & X.6 & Naskh & 9 & 185 & 2,573 & 10,666 & 67 & 0.63 & 51 & 192 & 0 \\
    \addlinespace[2pt]
    \cmidrule[0.25pt]{1-14}
    Ms16\us Oxford\us Bodl-Or-516 & 11th c., 1082 & 1 & III.3 & Naskh & 3 & 75 & 1,014 & 3,770 & 83 & 2.20 & 3 & 23 & 1 \\
    \addlinespace[2pt]
    \cmidrule[0.25pt]{1-14}
    \multirow{2}{2.9cm}{\raggedright Ms17\us Paris\us BnF-Arabe-6840} & \multirow{2}{1.8cm}{\raggedright 12th c., 1108} & \multirow{2}{*}{2} & X.6 & \multirow{2}{*}{Naskh} & 4 & 144 & 3,164 & 12,819 & 1 & 0.01 & 53 & 201 & 0 \\
     & & & X.12 & & 2 & 68 & 1,393 & 5,953 & 0 & 0.00 & 0 & 0 & 5 \\
    \addlinespace[2pt]
    \cmidrule[0.25pt]{1-14}
    Ms18\us Rampur\us Raza-3700 & 17th c., 1625 & 1 & X.6 & Naskh & 6 & 164 & 2,718 & 11,317 & 141 & 1.25 & 51 & 198 & 0 \\
    \addlinespace[2pt]
    \cmidrule[0.25pt]{1-14}
    \textbf{Subtotal Tiers 1--2} & & \textbf{23} & & & \textbf{119} & \textbf{2,969} & \textbf{47,778} & \textbf{206,185} & \textbf{7,740} & \textbf{3.75} & \textbf{686} & \textbf{2,616} & \textbf{35} \\
    \midrule
    \multicolumn{14}{l}{\textit{Tier 3: other scientific treatises (BnF, Gallica)}} \\
    \addlinespace[2pt]
    Ms\us Abu-l-wafa\us Paris\us BnF-Arabe-2494\us I.8 & 12th & 1 & I.8 & Naskh & 6 & 110 & 1,839 & 6,655 & 94 & 1.41 & 0 & 0 & 0 \\
    \addlinespace[2pt]
    \cmidrule[0.25pt]{1-14}
    Ms\us IbnalHaytham\us Paris\us BnF-Arabe-2458 & 1144 & 1 & -- & Naskh & 3 & 82 & 1,459 & 5,192 & 157 & 3.02 & 0 & 0 & 6 \\
    \addlinespace[2pt]
    \cmidrule[0.25pt]{1-14}
    Ms\us Farghani\us Paris\us BnF-Arabe-2504 & 17th--18th c. & 1 & -- & Naskh & 6 & 124 & 1,259 & 5,654 & 2 & 0.04 & 0 & 0 & 0 \\
    \addlinespace[2pt]
    \cmidrule[0.25pt]{1-14}
    Ms\us Farghani\us Paris\us BnF-Arabe-2546 & 1601 & 1 & -- & Naskh & 6 & 127 & 1,061 & 4,826 & 2 & 0.04 & 0 & 0 & 5 \\
    \addlinespace[2pt]
    \cmidrule[0.25pt]{1-14}
    Ms\us Biruni\us Paris\us BnF-Arabe-2497 & 17th c. & 1 & -- & Naskh & 4 & 82 & 625 & 2,955 & 33 & 1.12 & 0 & 0 & 0 \\
    \addlinespace[2pt]
    \cmidrule[0.25pt]{1-14}
    Ms\us Tusi\us Paris\us BnF-Arabe-2509\us II.12 & 1389 & 1 & II.12 & Naskh & 4 & 60 & 609 & 3,285 & 494 & 15.04 & 0 & 0 & 0 \\
    \addlinespace[2pt]
    \cmidrule[0.25pt]{1-14}
    Ms\us Jaghmini\us Paris\us BnF-Arabe-2500\us I.1 & 1404 & 1 & I.1 & Naskh & 5 & 68 & 743 & 5,204 & 1,844 & 35.43 & 2 & 2 & 2 \\
    \addlinespace[2pt]
    \cmidrule[0.25pt]{1-14}
    Ms\us IbnSina\us Paris\us BnF-Arabe-2484\us 1 & 1284 & 1 & 1 & Naskh & 3 & 72 & 848 & 3,629 & 0 & 0.00 & 0 & 0 & 6 \\
    \addlinespace[2pt]
    \cmidrule[0.25pt]{1-14}
    Ms\us Sizji\us Paris\us BnF-Arabe-4821 & 1149 & 1 & -- & Maghrebi & 13 & 199 & 1,934 & 6,895 & 137 & 1.99 & 0 & 0 & 0 \\
    \addlinespace[2pt]
    \cmidrule[0.25pt]{1-14}
    \textbf{Subtotal Tier 3} & & \textbf{9} & & & \textbf{50} & \textbf{924} & \textbf{10,377} & \textbf{44,295} & \textbf{2,763} & \textbf{6.24} & \textbf{2} & \textbf{2} & \textbf{19} \\
    \midrule
    \textbf{Total} & & \textbf{32} & & & \textbf{169} & \textbf{3,893} & \textbf{58,155} & \textbf{250,480} & \textbf{10,503} & \textbf{4.19} & \textbf{688} & \textbf{2,618} & \textbf{54} \\
    \bottomrule
    \end{tabular}
    \caption{Statistics of the manuscripts, one row per manuscript. ``Chars'' excludes spaces; ``Diacr.\ marks'' counts
    diacritic marks and ``\% diacr.'' is their share of ``Chars''. ``Numeral strings'' counts the numerals (transcribed
    or tagged) and ``Digits'' the digits transcribed. Manuscript identifiers match the dataset folder names to facilitate reproducibility.
    }
    \label{tab:manuscripts_stats}
\end{table*}

\section{Manual Annotation}
\label{sec:manual_annotation}
\subsection{Excerpt from the \textsc{Athaar} Annotation Guide}
\label{sec:excerpt_guide}
The corpus was annotated following a dedicated annotation guide
following historical manuscripts text editing conventions. Table~\ref{tab:manuscripts_stats}
reports the resulting editorial markers. The excerpt below covers
the transcription rules; the full guide will be made available upon acceptance together with  the
dataset repository.

\begin{figure}[t]
    \centering
    \includegraphics[width=\columnwidth]{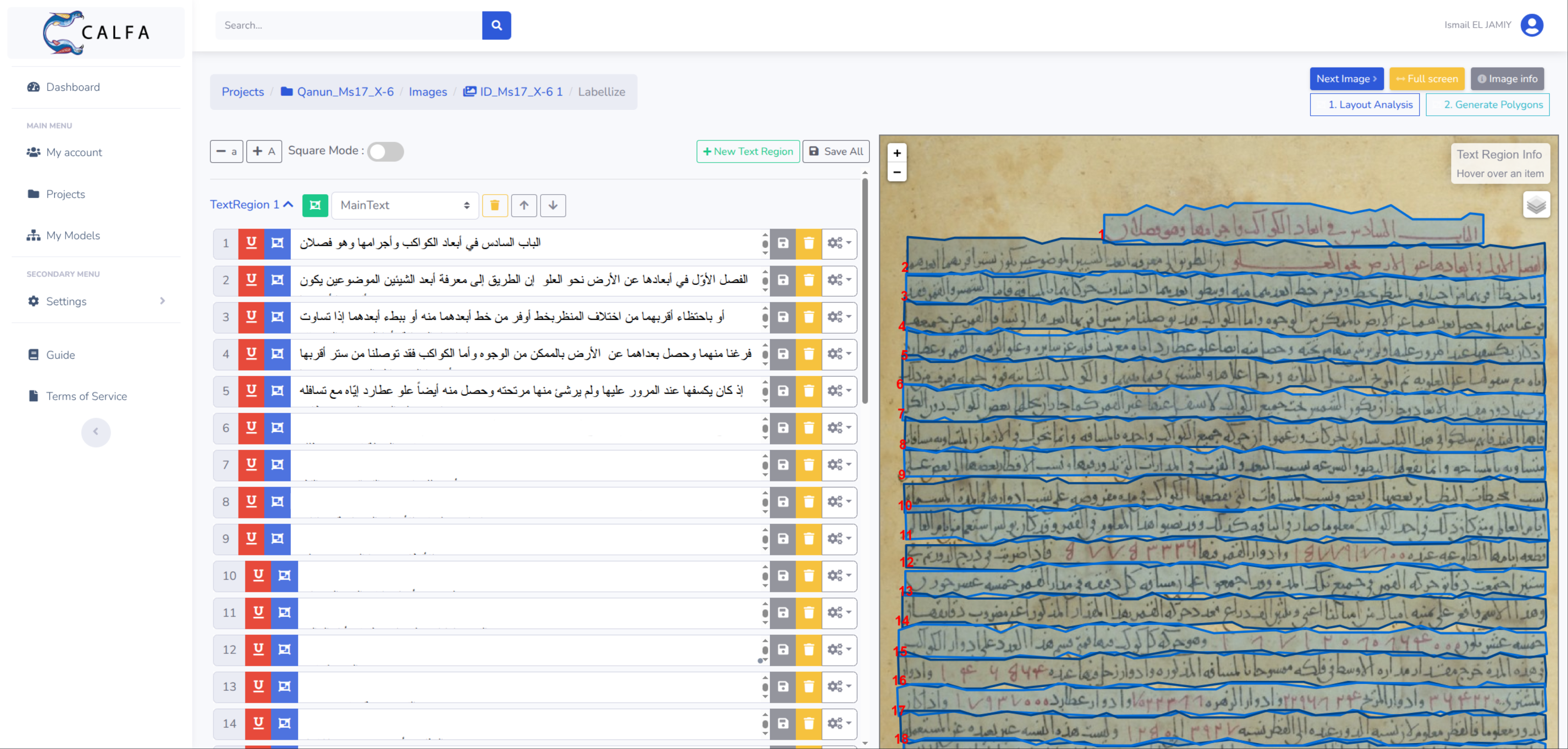}
    \caption{\textit{Calfa Vision} annotation interface.}
    \label{fig:calfa_platform}
\end{figure}
\subsubsection{Transcription Guidelines} \label{t-guidelines}

\paragraph{Core principle.} Transcribe \textbf{strictly what the scribe wrote}, without editorial emendation, modernization of orthography, or semantic interpretation. The transcription is diplomatic.

\paragraph{Main text scope.}

\begin{itemize}[nosep]
  \item \textbf{Transcribe:} Titles, section headings, and the running body text.
  \item \textbf{Exclude} (delete corresponding auto-detected regions):
  \begin{itemize}[nosep]
    \item Running headers
    \item Catchwords (page footers)
    \item Folio / page numbers
    \item Independent marginalia (glosses, corrections, external notes)
    \item Data tables
    \item Geometric diagrams and figures
  \end{itemize}
\end{itemize}

\begin{figure}[h]
\centering
\includegraphics[width=\columnwidth]{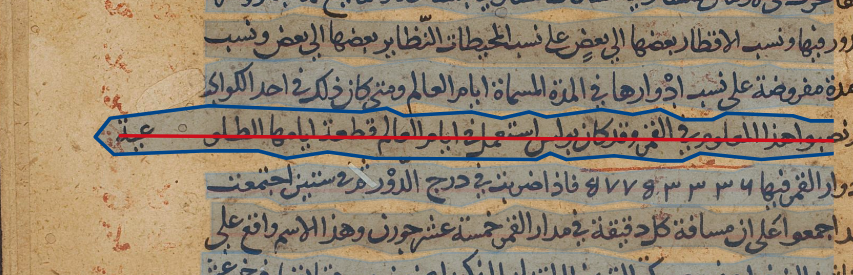}
\caption{Example of a word split across the text column and the margin, alongside expected transcription.}
\label{texte_marginal}
\end{figure}

\paragraph{Split words overflowing into margins.}
When a word starting in the main line overflows into the margin due to lack of space (see Figure ~\ref{texte_marginal}), transcribe the fragments as a single unified word on that line. Extend the baseline and polygon to encompass the marginal fragment.

\begin{warnbox}
Distinguish this from independent marginal annotations (glosses, reader notes, collations), which must \textbf{never} be transcribed.
\end{warnbox}

\begin{figure}
    \centering
    \includegraphics[width=\columnwidth]{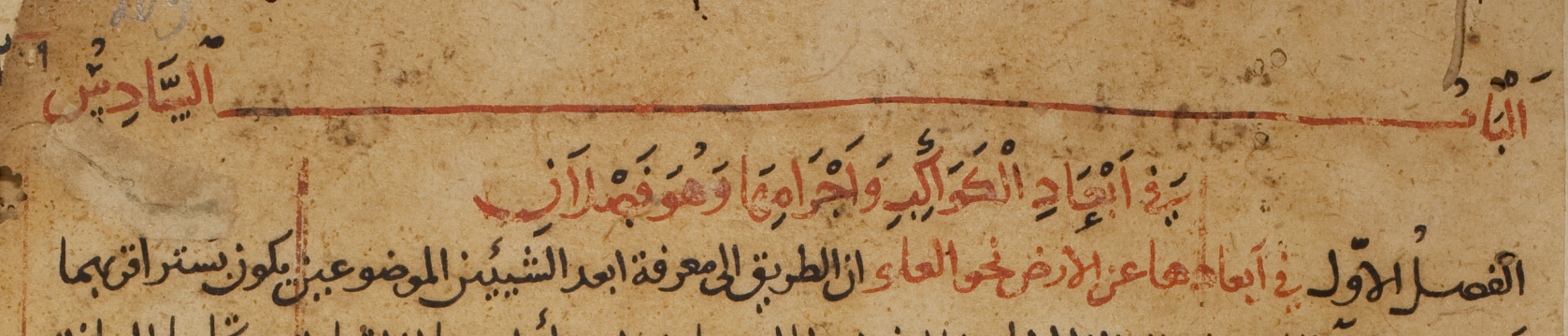}
    \caption{Example of partially vocalized text alongside unvocalized passages.}
    \label{fig:tachkil_partiel}
\end{figure}

\paragraph{Vocalization (Tashk\={\i}l).}
Historical texts are largely unvocalized. \textbf{Transcribe diacritics and short vowels only where explicitly written by the scribe. }Never add unwritten diacritics (see Figure~\ref{fig:tachkil_partiel}).

\begin{figure}
    \centering
    \includegraphics[width=0.7\columnwidth]{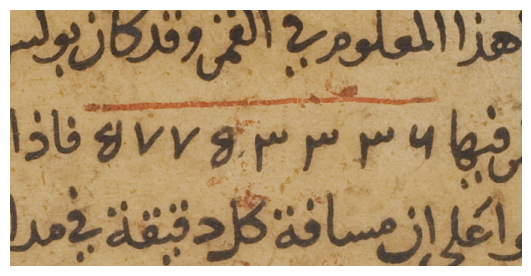}
    \caption{Overlined numerical digits}
    \label{fig:chiffres}
\end{figure}

\paragraph{Numerals.}
Scientific manuscripts frequently include numerical data, often overlined (see Figure~\ref{fig:chiffres}).

\textbf{Rules:}
\begin{itemize}[nosep]
  \item \textbf{If readable:} Transcribe the digits directly.
  \item \textbf{If ambiguous or difficult to read:} Insert a tag identifying the numeral system, as shown below:
\end{itemize}

\begin{center}
\small
\begin{tabularx}{\columnwidth}{Xl}
\toprule
\textbf{Case} & \textbf{Tag} \\
\midrule
Abjad numerals (alphanumerical notation) & \texttt{[abjad\_numbers]}) \\
Eastern Arabic-Indic numerals  & \texttt{[indian\_numbers]}) \\
Western Arabic numerals (0123456789) & \texttt{[arabic\_numbers]}) \\
Unidentified numeral system & \texttt{[unknown\_numbers]}) \\
\bottomrule
\end{tabularx}
\end{center}

\paragraph{Overlined elements.}
Non-numerical elements carrying a horizontal overline are tagged according to their function:

\begin{center}
\small
\begin{tabularx}{\columnwidth}{Xl}

\toprule
\textbf{Overlined Entity} & \textbf{Tag} \\
\midrule
Geometric label or object & \texttt{[geometrical\_object]}) \\
\bottomrule
\end{tabularx}
\end{center}

\paragraph{Illegible or damaged text.}
Where text is unreadable due to ink flaking, stains, holes, or physical decay:

\begin{itemize}[nosep]
  \item Insert \texttt{[unreadable]}) at the corresponding point in the line.
  \item If an entire line is unreadable, record solely \texttt{[unreadable\_line]}) for that line.
\end{itemize}

\paragraph{Elements excluded from segmentation.}
Delete any automatic TextRegion covering:
\begin{itemize}[nosep]
  \item Geometric diagrams
  \item Astronomical tables (grids of numerical entries)
  \item Decorative illuminations and ornaments
  \item Folio / page stamps and pagination
  \item Library ownership stamps
\end{itemize}

\paragraph{Tag summary.}
The final list of tags that can be usd is summurized below.
\begin{center}
\small
\begin{tabularx}{\columnwidth}{lX}
\toprule
\textbf{Tag} & \textbf{Description} \\
\midrule
\texttt{[abjad\_numbers]}) & Untranscribed Abjad numerals \\
\texttt{[indian\_numbers]}) & Untranscribed Eastern Arabic-Indic digits \\
\texttt{[arabic\_numbers]}) & Untranscribed Western Arabic digits \\
\texttt{[unknown\_numbers]}) & Numerals of unidentified notation \\
\texttt{[geometrical\_object]}) & Overlined geometric object notation \\
\texttt{[unreadable]}) & Illegible or damaged text within a line \\
\texttt{[unreadable\_line]}) & Completely illegible or damaged text line \\
\bottomrule
\end{tabularx}
\end{center}

\subsection{Annotation Challenges}
\label{sec:annotation_challenges}

The inter-annotators agreement results reported in Section~\ref{sec:quality}
show that even with a well-documented annotation guide, annotators still
disagreed on details that are inherently ambiguous in the manuscript images
rather than caused by annotator error. Two recurring sources of ambiguity
are illustrated below.

Figure~\ref{fig:indian_variant} shows a numeral string whose digit
forms do not match any standard numeral family: they
are variant forms of Eastern Arabic-Indic digits whose exact reading
is open to interpretation, even for an expert reader.

Figures~\ref{fig:missing_dots_1} and~\ref{fig:missing_dots_2} show
words in which the scribe omitted letter-distinguishing dots.
Since several Arabic letters share the same base form and differ only
by their dots, their absence makes the correct reading genuinely
uncertain without recourse to the surrounding context.

\begin{figure}[t]
    \centering
    \begin{subfigure}{\columnwidth}
        \includegraphics[width=\linewidth]{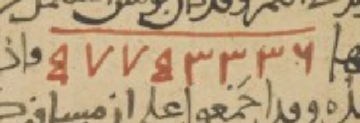}
        \caption{Variant Eastern Arabic-Indic digit forms whose reading
        is ambiguous.}
        \label{fig:indian_variant}
    \end{subfigure}
    \vspace{4pt}
    \begin{subfigure}{0.48\columnwidth}
        \includegraphics[width=\linewidth]{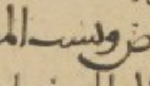}
        \caption{Letters with missing distinguishing dots (example 1).}
        \label{fig:missing_dots_1}
    \end{subfigure}
    \hfill
    \begin{subfigure}{0.48\columnwidth}
        \includegraphics[width=\linewidth]{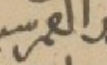}
        \caption{Letters with missing distinguishing dots (example 2).}
        \label{fig:missing_dots_2}
    \end{subfigure}
    \caption{Examples of annotation challenges arising from manuscript
    ambiguity rather than annotator error.}
    \label{fig:annotation_challenges}
\end{figure}

\section{Experimental details}
\label{sec:experiment_extra}

\subsection{Experiments Split Statistics}
\label{sec:splits}

Table~\ref{tab:experiment_sizes} reports the number of lines available for each experiment after the 70/30 query/support split described in Section~\ref{sec:experiments}.

\begin{table}[h]
\centering
\small
\setlength{\tabcolsep}{5pt}
\renewcommand{\arraystretch}{1.1}
\begin{tabular}{lccc}
\toprule
\textbf{Exper.} & \textbf{Total lines} & \textbf{Query (70\%)} & \textbf{Support (30\%)} \\
\midrule
\textsc{C-Scr} & 2{,}164 & 1{,}516 & 648 \\
\textsc{C-Ch} &   804   &   564   & 240 \\
\textsc{C-Tr} &   924   &   646   & 278 \\
\bottomrule
\end{tabular}
\caption{Line counts per experiment after the 70/30 query/support split.}
\label{tab:experiment_sizes}
\end{table}

\subsection{Hardware and Parameters Configfuration}
All models are run with temperature
$T = 0$ and \texttt{max\_tokens} $= 256$. Open-weight models are run on
a single NVIDIA RTX~8000 GPU (48\,GB VRAM) using Python~3.12 and
CUDA~12.8.

\subsection{Per-Witness Results}
Tables~\ref{tab:e1_detail}--\ref{tab:e3_detail} report per-witness CER and WER for each experiment.

\begin{itemize}
  \item \textbf{Experiment \textsc{C-Scr}} (Table~\ref{tab:e1_detail}):
  witnesses are divided into two groups according to the query script. The first group lists Naskh query witnesses, whose support exemplars are drawn from the Nasta'liq pool; the second lists Nasta'liq query witnesses, whose support is drawn from the Naskh pool.

  \item \textbf{Experiment \textsc{C-Ch}} (Table~\ref{tab:e2_detail}):
  each query witness belongs to Tier~2, i.e.\ a chapter of
  \textit{al-Q\=an\=un} other than X.6; its support exemplars
  are drawn from the Tier~1 witness of the same manuscript.

  \item \textbf{Experiment \textsc{C-Tr}} (Table~\ref{tab:e3_detail}):
  each query witness belongs to Tier~3, i.e.\ a treatise other than
  \textit{al-Q\=an\=un}; all queries share the same support bank,
  drawn once from all Tier~1 and Tier~2 witnesses.
\end{itemize}

\begin{table}[p]
\centering
\small
\setlength{\tabcolsep}{3pt}
\renewcommand{\arraystretch}{1.1}
\begin{tabularx}{\columnwidth}{Xcccccc}
\toprule
& \multicolumn{2}{c}{\textbf{Gemini}}
& \multicolumn{2}{c}{\textbf{CHURRO}}
& \multicolumn{2}{c}{\textbf{Qari-OCR}} \\
\cmidrule(lr){2-3}\cmidrule(lr){4-5}\cmidrule(lr){6-7}
\textbf{Witness} & CER & WER & CER & WER & CER & WER \\
\midrule
\multicolumn{7}{l}{\textit{Naskh queries — support from Nasta'liq pool}} \\
\midrule
Ms01 (Berlin 1613)    & 14.3 &  48.4 &  64.2 & 115.2 & 175.8 & 225.0 \\
Ms04 (Cairo 866)      & 34.1 & 123.9 &  62.2 & 109.8 & 106.5 & 153.3 \\
Ms06 (Istanbul 87)    & 19.6 &  86.7 &  54.5 &  93.9 & 123.5 & 155.1 \\
Ms07 (Konya 6625)     & 13.4 &  41.5 &  74.6 & 125.2 & 247.9 & 338.2 \\
Ms13 (BL Or-1997)     & 19.8 &  59.0 & 110.9 & 153.4 & 152.6 & 188.2 \\
Ms14 (Mashhad 5588)   & 12.6 &  47.5 &  73.8 & 116.7 & 160.9 & 201.0 \\
Ms15 (Mashhad 12292)  & 19.2 &  61.0 &  64.3 & 125.2 & 200.6 & 274.8 \\
Ms17 (BnF Ar. 6840)   &  8.1 &  32.2 &  59.5 & 101.0 & 132.3 & 171.8 \\
Ms18 (Rampur 3700)    &  7.5 &  28.5 &  64.1 & 112.1 & 120.7 & 173.9 \\
\midrule
\textbf{\textsc{C-Scr}(N) }  & \textbf{16.3} & \textbf{62.3} & \textbf{70.0} & \textbf{115.6} & \textbf{154.5} & \textbf{198.9} \\
\midrule
\multicolumn{7}{l}{\textit{Nasta'liq queries — support from Naskh pool}} \\
\midrule
Ms02 (Berlin 275)      & 22.0 & 67.1 &  65.3 & 113.3 & 233.2 & 291.6 \\
Ms05 (Hyderabad 374)   & 15.6 & 48.0 & 134.4 & 190.0 & 161.4 & 242.0 \\
Ms08 (Istanbul 1498)   & 27.2 & 72.9 &  77.0 & 140.0 & 147.3 & 215.5 \\
Ms10 (Istanbul 65)     & 11.2 & 33.2 &  49.7 & 104.1 & 110.6 & 151.8 \\
\midrule
\textbf{\textsc{C-Scr}(NT) } & \textbf{18.5} & \textbf{53.8} & \textbf{79.9} & \textbf{134.9} & \textbf{158.6} & \textbf{219.5} \\
\bottomrule
\end{tabularx}
\caption{Per-witness CER and WER (\%) for \textsc{C-Scr} at $k=3$ (cross-script).}
\label{tab:e1_detail}
\end{table}

\begin{table}[p]
\centering
\small
\setlength{\tabcolsep}{3pt}
\renewcommand{\arraystretch}{1.1}
\begin{tabularx}{\columnwidth}{Xcccccc}
\toprule
& \multicolumn{2}{c}{\textbf{Gemini}}
& \multicolumn{2}{c}{\textbf{CHURRO}}
& \multicolumn{2}{c}{\textbf{Qari-OCR}} \\
\cmidrule(lr){2-3}\cmidrule(lr){4-5}\cmidrule(lr){6-7}
\textbf{Witness} & CER & WER & CER & WER & CER & WER \\
\midrule
Ms01 (Berlin 1613)    &  5.6 & 26.9 & 35.3 &  85.0 & 156.4 & 171.8 \\
Ms02 (Berlin 275)     & 25.3 & 60.3 & 55.5 & 109.3 & 250.2 & 343.0 \\
Ms04 (Cairo 866)      & 28.0 & 99.7 & 40.7 &  91.0 & 124.5 & 117.5 \\
Ms06 (Istanbul 87)    & 23.3 & 82.9 & 57.1 & 108.0 & 203.3 & 168.3 \\
Ms08 (Istanbul 1498)  & 35.0 & 86.0 & 111.0 & 167.5 & 106.4 & 172.0 \\
Ms10 (Istanbul 65)    & 11.0 & 32.5 & 62.1 & 108.9 & 145.8 & 219.2 \\
Ms13 (BL Or-1997)     & 15.3 & 56.9 & 69.6 & 123.1 & 107.0 & 142.5 \\
Ms17 (BnF Ar. 6840)   & 15.2 & 43.9 & 56.0 & 113.1 & 216.0 & 298.5 \\
\midrule
\textbf{\textsc{C-Ch} } & \textbf{18.9} & \textbf{63.2} & \textbf{59.3} & \textbf{109.2} & \textbf{163.1} & \textbf{194.0} \\
\bottomrule
\end{tabularx}
\caption{Per-witness CER and WER (\%) for \textsc{C-Ch} at $k=3$ (cross-chapter transfer).}
\label{tab:e2_detail}
\end{table}

\begin{table}[p]
\centering
\small
\setlength{\tabcolsep}{3pt}
\renewcommand{\arraystretch}{1.1}
\begin{tabularx}{\columnwidth}{Xcccccc}
\toprule
& \multicolumn{2}{c}{\textbf{Gemini}}
& \multicolumn{2}{c}{\textbf{CHURRO}}
& \multicolumn{2}{c}{\textbf{Qari-OCR}} \\
\cmidrule(lr){2-3}\cmidrule(lr){4-5}\cmidrule(lr){6-7}
\textbf{Witness} & CER & WER & CER & WER & CER & WER \\
\midrule
Ms Ar 2494 (Ab\={u} al-Waf\={a})  & 25.1 &  63.3 &  68.7 & 105.8 & 153.7 & 175.2 \\
Ms Ar 2458 (Ibn al-Haytham)        & 29.3 &  73.2 &  75.3 & 114.7 & 335.7 & 335.8 \\
Ms Ar 2504 (Fargh\={a}n\={i})      &  2.3 &  13.7 &  38.4 & 107.4 & 260.6 & 381.1 \\
Ms Ar 2546 (Fargh\={a}n\={i})      & 13.9 &  39.8 &  49.3 & 110.8 & 217.5 & 319.4 \\
Ms Ar 2497 (B\={\i}r\={u}n\={i})   &  7.5 &  25.9 &  32.8 & 104.9 & 275.6 & 377.8 \\
Ms Ar 2509 (T\={u}s\={i})          & 16.3 &  78.7 &  37.8 &  93.1 & 252.9 & 282.2 \\
Ms Ar 2500 (Jaghm\={\i}n\={\i})    & 20.9 &  60.6 &  63.1 &  98.8 & 195.3 & 137.6 \\
Ms Ar 2484 (Ibn S\={\i}n\={a})     & 29.2 &  79.3 &  77.6 & 136.2 & 309.1 & 375.0 \\
Ms Ar 4821 (Sijz\={\i})            & 29.9 &  75.0 & 136.4 & 181.5 & 344.3 & 279.6 \\
\midrule
\textbf{\textsc{C-Tr} } & \textbf{20.4} & \textbf{61.1} & \textbf{63.9} & \textbf{112.5} & \textbf{252.9} & \textbf{260.8} \\
\bottomrule
\end{tabularx}
\caption{Per-witness CER and WER (\%) for \textsc{C-Tr} at $k=3$ (cross-treatise transfer).}
\label{tab:e3_detail}
\end{table}

\subsection{Ablation Study: Number of Exemplars} 
\label{sec:k_ablation}
To select $k$, we ran a preliminary ablation with Gemini~3~Flash on
Tier~1, using five witnesses as the support pool and eight as the query
pool. For each $k \in \{1, 3, 5, 7, 10\}$, three trials were run with independently drawn support lines and scored with CER and WER.
Results are reported in Table~\ref{tab:kshot_ablation}. No condition outperforms the others by a margin exceeding the trial-to-trial standard deviation, suggesting the model reaches its performance ceiling regardless of $k$ on this task. 

We therefore set \textbf{$k = 3$} for all experiments, since no value of $k$ outperforms the others by a statistically meaningful margin.

\begin{table}[p]
\centering
\small
\setlength{\tabcolsep}{5pt}
\renewcommand{\arraystretch}{1.1}
\begin{tabular}{lll}
\toprule
$k$ & \textbf{CER (\%)}$\downarrow$ & \textbf{WER (\%)}$\downarrow$ \\
\midrule
1  & $14.08 \pm 0.48$ & $38.79 \pm 1.95$ \\
3  & $14.55 \pm 0.90$ & $39.73 \pm 0.81$ \\
5  & $14.31 \pm 1.28$ & $38.66 \pm 1.94$ \\
7  & $15.11 \pm 0.21$ & $39.63 \pm 0.72$ \\
10 & $14.39 \pm 1.59$ & $38.64 \pm 2.97$ \\
\bottomrule
\end{tabular}
\caption{$k$-shot ablation on Tier~1 (Gemini~3~Flash, 3 trials per $k$). Scores are reported in CER and WER.}
\label{tab:kshot_ablation}
\end{table}

\subsection{Few-shot Prompt}

\begin{promptbox}{Few-shot prompt}
You transcribe single lines from historical Arabic scientific manuscripts.

\textbf{Task}: write exactly what the scribe wrote on the line in the TEST IMAGE.

\begin{enumerate}
\item Do not correct, normalize or modernize it.
\item Write diacritics (vowel signs, shadda, sukun) only where they are visible. Never add them.
\item Write hamza and letter dots only as they appear in the image.
\item Write numerals with Eastern Arabic-Indic digits (e.g.\ \texttt{0--9} rendered in their Arabic-Indic form).
\item If a word is illegible, skip it. Never guess or complete it from context.
\item Output only the transcription of the TEST IMAGE, on one line: no translation, explanation, quotes or formatting.
\end{enumerate}

The EXAMPLES below show the expected output format. Do not copy their text into your answer.

\textbf{Examples}:

\texttt{<example>}
\texttt{<image>} [support image $1$] \texttt{</image>}
\texttt{<transcription>} [support transcription $1$] \texttt{</transcription>}
\texttt{</example>}

$\vdots$

\texttt{<example>}
\texttt{<image>} [support image $s$] \texttt{</image>}
\texttt{<transcription>} [support transcription $s$] \texttt{</transcription>}
\texttt{</example>}

Test Image:
\texttt{<image>} [query image] \texttt{</image>}
\texttt{<transcription>}
\end{promptbox}

\end{document}